\documentclass[11pt]{article}

\usepackage[preprint]{acl}

\usepackage{times}
\usepackage{latexsym}

\usepackage[T1]{fontenc}

\usepackage[utf8]{inputenc}

\usepackage{microtype}

\usepackage{inconsolata}

\usepackage{graphicx}
\usepackage{booktabs}
\usepackage{multirow}
\usepackage{amsmath}
\usepackage{algorithm}
\usepackage{algorithmic}
\usepackage{tcolorbox}
\tcbuselibrary{listings,breakable}

\newcommand{\method}{SkillComposer}

\title{SkillComposer: Learning to Evolve Agent Skills for Specification and Generalization}

\author{
 \textbf{Qi Zhang\textsuperscript{1,2}\thanks{Equal contribution.}\thanks{Correspondence to: \texttt{cheung\_se@zju.edu.cn}, \texttt{xuanjie.wxb@alibaba.com}}},
 \textbf{Zhaopeng Feng\textsuperscript{2,3}\footnotemark[1]},
 \textbf{Xiaonan Shi\textsuperscript{2}},
 \textbf{Xiaomeng Hu\textsuperscript{1}},
\\
 \textbf{Chu Liu\textsuperscript{2}},
 \textbf{Pengjun Xie\textsuperscript{2}},
 \textbf{Xiaobin Wang\textsuperscript{2}\thanks{Corresponding authors.}},
 \textbf{Jieping Ye\textsuperscript{2}},
\\
 \textbf{Bryan Hooi\textsuperscript{3}},
 \textbf{Haobo Wang\textsuperscript{1}},
 \textbf{Junbo Zhao\textsuperscript{1}\footnotemark[3]}
\\
\\
 \textsuperscript{1}Zhejiang University,
 \textsuperscript{2}Tongyi Lab,
 \textsuperscript{3}National University of Singapore
}

\begin{document}
\maketitle
\begin{abstract}
Agent skills, which consist of reusable strategies that guide agent reasoning and action, have shown strong potential for improving model capability at inference time. However, current skill construction methods treat the problem as one-shot extraction, overlooking a fundamental tension: a skill tailored to the specific task fails to transfer, while the abstracted skill often provides insufficient guidance. We attribute this fragility to the absence of explicit mechanisms for skill \textit{specification} and \textit{generalization}. To address this gap, we introduce \method{}, a framework that decomposes skill construction into three learnable operations: \textit{create}, \textit{improve}, and \textit{merge}. Trained via systematic rejection sampling recipe, \method{} enables language models to self-evolve skills at inference time and supports three deployment modes: offline for building generalized libraries, online for task-specific refinement, and hybrid for combining both. Comprehensive experiments on $\tau^2$-Bench, LiveCodeBench v6, and AppWorld show that \method{} consistently outperforms baselines. Our SkillComposer-4B improves a 27B executor by up to +4.5 on agent tasks and +3.4 on code tasks, while generalizing across domains and task types unseen during training. Analysis reveals that \textit{merge} and \textit{improve} address orthogonal quality dimensions and that skill composition is a transferable meta-ability, providing a practical recipe for skill-augmented inference.

\end{abstract}

\section{Introduction}

Large language models (LLMs) have demonstrated remarkable progress across a wide range of tasks, yet they still struggle with novel scenarios that demand structured problem-solving strategies~\cite{agent_skills_survey}. To bridge this gap, recent work has introduced \textit{agent skills}: reusable natural-language instructions, each comprising a name, a trigger condition, and a procedural body, that encode domain-specific procedures in a modular format~\cite{anthropic_skills,programmatic_skill}. By loading relevant skills into the context, models can leverage accumulated experience without parameter updates, offering a lightweight and interpretable mechanism for capability enhancement~\cite{skill_net}.

\begin{figure}[t]
    \centering
    \includegraphics[width=\columnwidth]{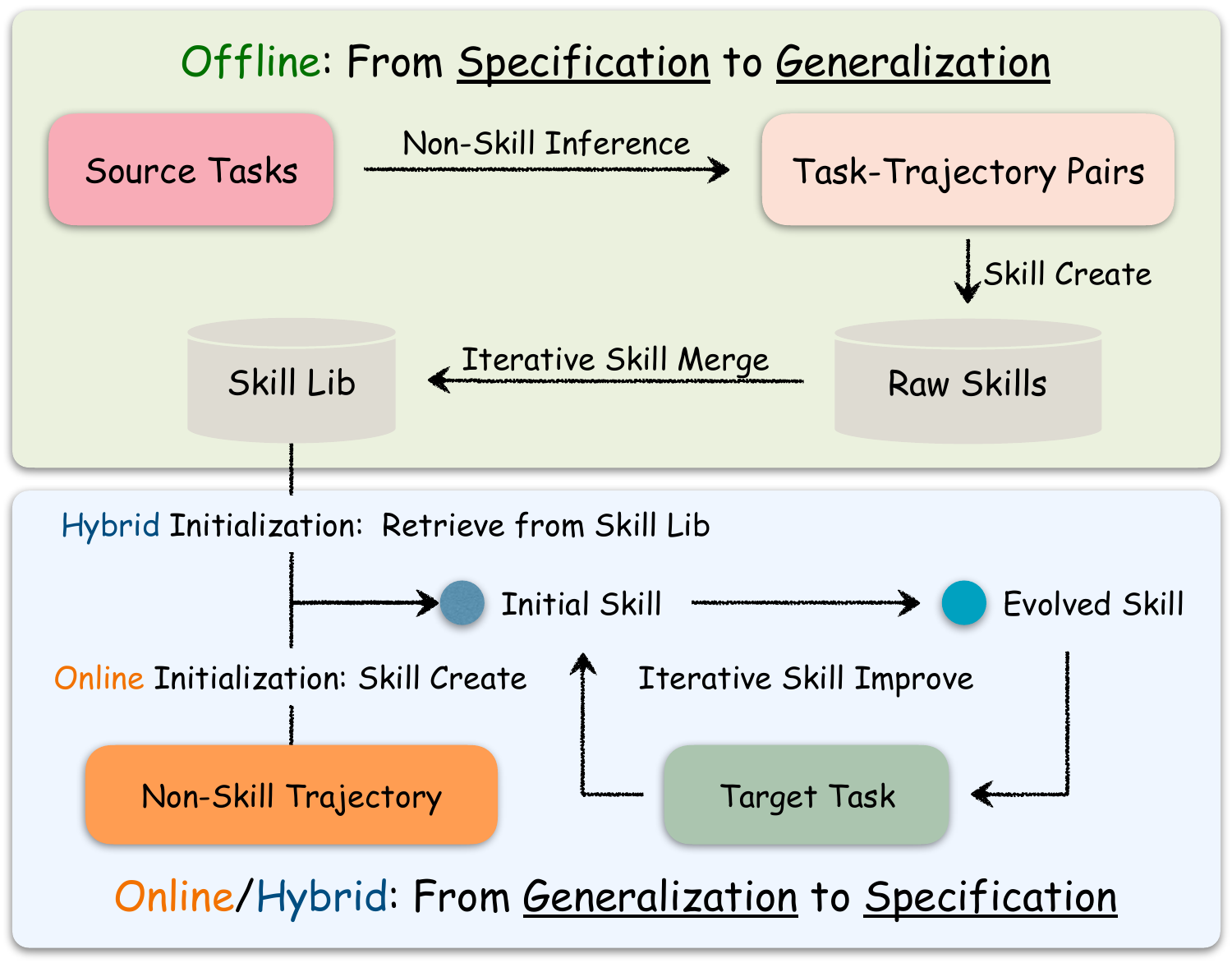}
    \caption{Three deployment modes of \method{}. \textit{Offline} builds a generalized skill library via creation and merging; \textit{Online} creates and iteratively refines skills per task; \textit{Hybrid} initializes from an offline library and specializes via online evolution.}
    \label{fig:online-offline}
\end{figure}

Despite their promise, current skill-based approaches face several fundamental limitations: they rely on high-quality human-written skills~\cite{skillsbench} or on extracting procedural knowledge from successful trajectories~\cite{memp}, both of which are costly and difficult to scale. When models generate skills autonomously, the results often yield limited or even negative performance gains, as the models lack mechanisms for quality control, refinement, and abstraction~\cite{trace2skill,agent_skills_analysis}. More critically, effective skills require both \textit{generalization} (abstracting across similar tasks to produce transferable strategies) and \textit{specification} (tailoring skills to particular task patterns), yet existing methods provide no systematic way to achieve either. A skill that is too specific fails to transfer to other relevant tasks, while one that is too general provides insufficient guidance for other particular tasks~\cite{agent_skills_analysis, agent_skills_github}.

In this work, we observe that skill construction can be decomposed into three learnable abilities: (1) \textbf{Skill Create} extracts reusable procedural knowledge from raw execution trajectories, (2) \textbf{Skill Merge} drives \textit{generalization} by consolidating semantically similar skill pairs into broader, more transferable ones, and (3) \textbf{Skill Improve} drives \textit{specification} by refining a skill to better capture task-specific patterns based on new execution experience. Together, these operations define a complete lifecycle for skill evolution: from initial extraction, through cross-task abstraction to iterative refinement. We propose \textbf{\method{}}, which enables language models to self-evolve its skills at inference time \textbf{without ground-truth supervision}. Moreover, we use delta pass rate guided rejection sampling to synthesize supervised fine-tuning data to further enhance these three abilities.

Building on these abilities, \method{} supports three modes of operation that address different practical scenarios (Figure~\ref{fig:online-offline}). In \textit{Offline} mode, the model builds a general-purpose skill library from a training set via iterative creation and merging, and retrieves relevant skills for new tasks at inference time. In \textit{Online} mode, the model starts with an empty library and iteratively creates and refines skills on the fly, driving specification in real time. This mode requires no prior data and is applicable to open-ended or streaming scenarios. \textit{Hybrid} mode combines both by initializing with a generalized offline library and then specializing it through online evolution, balancing broad coverage with task-specific adaptation. We evaluate \method{} on code generation (LiveCodeBench v6) and agent tasks ($\tau^2$-Bench, AppWorld), demonstrating consistent improvements across models and domains. Our contributions are as follows:
\begin{itemize}
    \item We identify specification and generalization as two orthogonal dimensions of skill quality, and decomposing skill construction into creating, improving, and merging provides an explicit mechanism for this trade-off.
    \item We propose \method{}, a rejection-sampling-based framework that teaches language models to compose, refine, and abstract skills, supporting offline, online, and hybrid deployment modes.
    \item Comprehensive experiments across multiple benchmarks, model scales, and both in-domain and out-of-distribution settings demonstrate the effectiveness and generalization of \method{}. Analysis further reveals that Skill Merge and Improve address complementary quality dimensions and that skill composition transfers across task types, providing a practical recipe for skill-augmented inference.
\end{itemize}

\section{Related Work}
\noindent\textbf{Agent skills and skill libraries.}
Agent skills have recently emerged as a practical abstraction for extending LLM agents at inference time. Unlike a single tool call, a skill package reusable procedural knowledge, trigger conditions, and sometimes executable resources that can be loaded on demand~\cite{anthropic_skills,agent_skills_survey}. Recent work has therefore studied how to build, organize, and reuse skill libraries. SkillX constructs a plug-and-play skill knowledge base by distilling trajectories into hierarchical skills and refining them with execution feedback~\cite{skillx}. SkillRL distills past experience into a hierarchical SkillBank and lets the skill library co-evolve with an agent policy during reinforcement learning~\cite{skillrl}. Complementary work upgrades textual skills into executable program functions that intervene in the agent loop~\cite{hasp}. These works show that skills can serve as compact experience carriers for LLM agents. However, they mainly treat skills as external artifacts to be collected, retrieved, or optimized, while the ability to construct high-quality skills remains dependent on human authoring, successful trajectories, or external optimization loops.

\begin{figure*}[t]
    \centering
    \includegraphics[width=\textwidth]{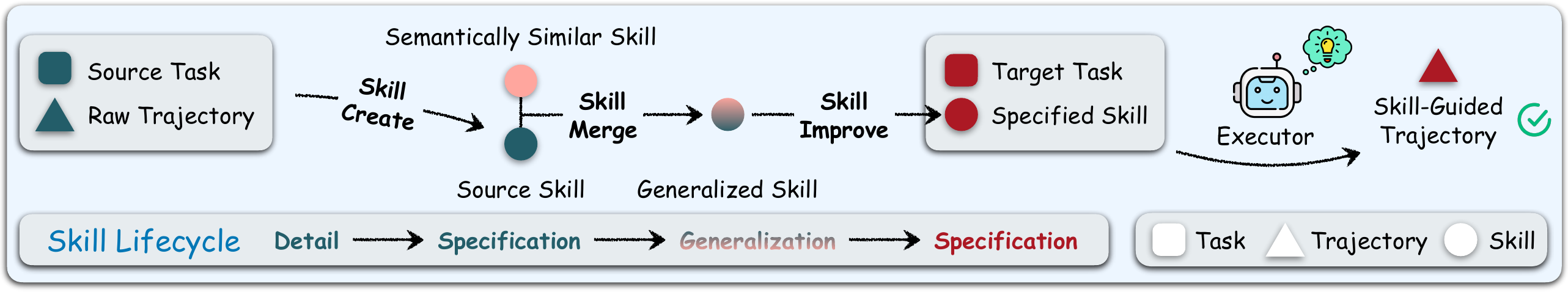}
    \caption{The lifecycle of a skill under \method{}. A skill is first created from a raw execution trajectory (\textit{creation}). Multiple semantically similar while task-specific skills are then consolidated via Skill Merge to form broader, more transferable skills (\textit{specification $\rightarrow$ generalization}). When applied to a new task, a generalized skill is iteratively refined via Skill Improve to better fit the target task (\textit{generalization $\rightarrow$ specification}). The resulting specialized skill is finally consumed by the skill executor to guide inference.}
    \label{fig:skill_lifecycle}
\end{figure*}

\noindent\textbf{Skill evolution and quality control.}
Closest to our work are recent methods that move from static skill reuse to automatic skill evolution. EvoSkill discovers and edits skills through iterative failure analysis and validation~\cite{evoskill}; CoEvoSkills couples a skill generator with a surrogate verifier to build multi-file skill packages without ground-truth test content~\cite{coevoskills}; and SkillClaw aggregates cross-user interaction traces to collectively refine a shared skill repository~\cite{skillclaw}. Meanwhile, benchmark studies reveal that skill benefits are fragile: curated skills can substantially improve agents, but self-generated skills may provide no average gain~\cite{skillsbench}, and skill utility degrades under realistic retrieval and adaptation settings~\cite{skillswild}. These findings suggest that the core bottleneck is not merely having more skills, but constructing skills with explicit quality control, task-level specification, and cross-task generalization. Different from prior work, \method{} decomposes skill construction into three trainable abilities--\textit{create}, \textit{improve}, and \textit{merge}--and learns them through rejection-sampled supervision filtered by delta pass rate. In this way, skill evolution becomes an intrinsic capability of the language model itself: Skill Improve drives specification, Skill Merge drives generalization, and the resulting model can self-evolve skill libraries in offline, online, and hybrid settings.

\section{Methodology}
In this section, we dive into the framework of \method{}. We first formalize three core skill operations (\S\ref{subsec:skill-ops}), then describe how they compose into three inference-time application scenarios (\S\ref{subsec:skill-evolution}), and finally introduce a rejection-sampling-based training procedure that strengthens these operations (\S\ref{subsec:rejection-sampling}). Figure~\ref{fig:skill_lifecycle} provides an overview of the lifecycle of a skill under our proposed \method{}.

\subsection{Preliminary}
We begin by formalizing the concept of agent skills. In the most general form, agent skills comprise a markdown-style document \texttt{SKILL.md} along with optional auxiliary files such as references, scripts, and assets~\cite{anthropic_skills}.
In this work, we focus on the main body of agent skills, where each skill $s$ is defined as a tuple of three components: a name $n$, a description $d$, and a body $b$. The name serves as an identifier, the description specifies the trigger condition under which the skill should be activated, and the body encodes the procedural strategy to be followed.

\subsection{Skill Operations Formulation}
\label{subsec:skill-ops}
Maintaining a high-quality skill library requires addressing three fundamental challenges: \textit{i}) how to extract a reusable skill from a raw trajectory produced without any skill guidance, \textit{ii}) how to prevent the unbounded growth of a skill library while promoting generalization, and \textit{iii}) how to refine an existing skill based on new execution experience. We address these challenges by introducing three composable operations: \textit{create}, \textit{merge}, and \textit{improve}.

\noindent \textit{\textbf{Skill Create.}} When the agent solves a task without leveraging any prior skill, the resulting trajectory contains implicit procedural knowledge that can benefit future tasks with a similar structure. Skill Create distills such a trajectory into a reusable skill:
\begin{equation}
\nonumber
  s={\rm SkillComposer_{create}}(x,{\rm LLM}(x))
\end{equation}

\noindent \textit{\textbf{Skill Merge.}} As skills accumulate, the library risks unbounded growth and over-specification, where many narrowly-scoped skills overlap in functionality. To mitigate this, Skill Merge consolidates two semantically similar skills, $s_1$ and $s_2$, into a single, more general one:
\begin{equation}
\nonumber
  s={\rm SkillComposer_{merge}}(s_1,s_2)
\end{equation}
Specifically, to measure the semantical similarity of two skills, we introduce a multi-view similarity mechanism where we compute similarity across every dimension for each skill pair:
\begin{equation}
\label{eq:multi-view}
  {\rm sim}(s_1,s_2)=\frac{1}{3}\sum_{I\in\{n,d,b\}}{\rm sim}(I_1,I_2)
\end{equation}
Only pairs with similarity larger than a threshold $\delta$ will be selected to conduct Skill Merge.

\noindent \textit{\textbf{Skill Improve.}} When the agent solves a task under the guidance of an existing skill $s_{\rm o}$, the resulting trajectory may reveal new patterns or refinements that the original skill does not capture. Skill Improve incorporates these insights to produce a more effective skill:
\begin{equation}
\nonumber
  s={\rm SkillComposer_{improve}}(x,s_{\rm o},{\rm LLM}(x,s_{\rm o}))
\end{equation}

\subsection{Inference-Time Skill Evolution}
\label{subsec:skill-evolution}
The three operations can be chained in different orders depending on whether a pre-existing task set is available and whether task-specific refinement is desired at inference time. We organize these compositions into three deployment modes.  Algorithm~\ref {alg:skillevolver} also unifies the three modes into a single procedure:

\noindent\textbf{Offline.} The most straightforward way to leverage skills at inference time is to pre-build a skill library from a given task set. In offline mode, we first generate raw trajectories for all tasks and apply Skill Create to extract one skill per trajectory. We then compute pairwise similarities using Equation~\ref{eq:multi-view} and iteratively merge the most similar pairs until no pair exceeds $\delta$. The resulting library captures generalized procedural knowledge. At inference time, we retrieve the top-$k$ skills for each target task and employ a self-select mechanism that allows the executor to choose the most appropriate skill, or abstain if none is relevant.

\noindent\textbf{Online.} In scenarios where no pre-existing task set is available, we operate in online mode. For each incoming task, the agent generates a raw trajectory and applies Skill Create to initialize a skill. The skill is then iteratively refined: at each iteration, the agent re-executes the task under the current skill's guidance, and Skill Improve updates the skill based on the new trajectory. This process repeats for a fixed number of iterations $T$, progressively driving specification.

\noindent\textbf{Hybrid.} The hybrid mode combines the strengths of both approaches. It initializes with an offline-constructed skill library, then applies online evolution on top. For each new task, we retrieve and self-select a relevant skill from the library, then refine it through iterative improvement. This enables the model to start from a generalized foundation and progressively specialize toward the target task.

\subsection{Training via Rejection Sampling}
\label{subsec:rejection-sampling}
While any language model can perform the three operations zero-shot, the output quality varies considerably. We strengthen them via supervised fine-tuning on rejection-sampled data~\cite{rejection_sampling}, using a unified quality signal: \textit{delta pass rate}, the performance change caused by a candidate skill. A training example is accepted only when the candidate improves the executor's pass@$k$ relative to the appropriate baseline by at least a threshold~$\epsilon$.

Concretely, we generate $n$ raw trajectories per task and apply the base model to produce candidate skills. For \textit{create}, we compare the executor's pass@$1$ with and without the candidate skill on the source task and retain the example if the improvement exceeds $\epsilon_{\rm c}$:
\begin{align}
  \Delta_{\text{create}} &= \text{pass@}1(\text{LLM}(x, s)) \notag \\
  &\quad - \text{pass@}1(\text{LLM}(x)) \geq \epsilon_{\rm c}
\end{align}
For \textit{merge} and \textit{improve}, the criterion generalizes to multiple evaluation tasks. Merge assesses the consolidated skill $s$ against the two original skills $s_1, s_2$ on their respective source tasks $x_1, x_2$:
\begin{align}
  \Delta_{\text{merge}} &= \frac{1}{2}\sum_{i=1}^{2} \big[\text{pass@}1(\text{LLM}(x_i, s)) \notag \\
  &\quad - \text{pass@}1(\text{LLM}(x_i, s_i))\big] \geq \epsilon_{\rm m}
\end{align}
Improve evaluates the refined skill against the original skill $s_{\rm o}$ on both the source task $x$ (from which $s_{\rm o}$ was derived) and a target task $x'$:
\begin{align}
  \Delta_{\text{improve}} &= \frac{1}{2}\sum_{x_i \in \{x, x'\}} \big[\text{pass@}1(\text{LLM}(x_i, s)) \notag \\
  &\quad - \text{pass@}1(\text{LLM}(x_i, s_{\rm o}))\big] \geq \epsilon_{\rm i}
\end{align}
For improve, we include both cross-task ($x' \neq x$) and same-task ($x' = x$) pairs, so the training signal captures both specification and generalization. Detailed algorithms for all three procedures are provided in Appendix~\ref{sec:rejection-sampling-appendix}.

\begin{algorithm}[!t]
\caption{Inference-Time Skill Evolution}
\label{alg:skillevolver}
\begin{algorithmic}[1]
\REQUIRE Task set $\mathcal{X}$, skill executor LLM, mode $\in$ \{offline, online, hybrid\}, max iterations $T$, skill library $\mathcal{L} \leftarrow \emptyset$
\ENSURE Evolved skill library $\mathcal{L}$
\STATE \textbf{Stage 1: Skill Library Initialization}
\IF{mode $\in$ \{offline, hybrid\}}
    \FOR{each task $x \in \mathcal{X}$}
        \STATE $\tau \leftarrow \text{LLM}(x)$ \hfill $\triangleright$ Raw trajectory
        \STATE $s \leftarrow \text{SkillComposer}_{\text{create}}(x, \tau)$
        \STATE $\mathcal{L} \leftarrow \mathcal{L} \cup \{s\}$
    \ENDFOR
    \REPEAT
        \STATE $(s_i, s_j) \leftarrow \text{RetrieveSimilarPair}(\mathcal{L},\delta)$
        \STATE $s \leftarrow \text{SkillComposer}_{\text{merge}}(s_i, s_j)$
        \STATE $\mathcal{L} \leftarrow (\mathcal{L} \setminus \{s_i, s_j\}) \cup \{s\}$
    \UNTIL{no similar pairs remain}
\ENDIF
\STATE \textbf{Stage 2: Skill Evolution}
\FOR{each target task $x'$}
    \IF{mode = offline}
        \STATE $s \leftarrow \text{SelfSelect}(\text{Retrieve}(\mathcal{L}, x', k))$
        \STATE $\tau \leftarrow \text{LLM}(x', s)$
    \ELSE
        \IF{mode = online}
            \STATE $\tau \leftarrow \text{LLM}(x')$
            \STATE $s \leftarrow \text{SkillComposer}_{\text{create}}(x', \tau)$
        \ELSIF{mode = hybrid}
            \STATE $s \leftarrow \text{SelfSelect}(\text{Retrieve}(\mathcal{L}, x', k))$
        \ENDIF
        \FOR{$t = 1$ to $T$}
            \STATE $\tau \leftarrow \text{LLM}(x', s)$
            \STATE $s \leftarrow \text{SkillComposer}_{\text{improve}}(x', s, \tau)$
        \ENDFOR
    \ENDIF
    \STATE $\mathcal{L} \leftarrow \mathcal{L} \cup \{s\}$
\ENDFOR
\end{algorithmic}
\end{algorithm}

\section{Experiments}
\subsection{Experimental Settings}
\paragraph{Training and Evaluation}
For coding task, we use OpenCodeReasoning~\cite{opencodereasoning} for both rejection sampling and building the initial skill library.
For rejection sampling of the agent task, we follow AutoForge~\citep{cai2025autoforge} to synthesize datasets similar to the Retail domain in $\tau^2$-Bench. 
In summary, we synthesize around 7,000 supervised fine-tuning data points. We further train Qwen3.5-4B~\cite{qwen35blog} with LlamaFactory, and run skill executor and SkillComposer using vllm.
We extend more implementation details in Appendix~\ref{app:implementation}.

\paragraph{Benchmarks}
We evaluate \method{} on three benchmarks spanning agent and code tasks. (1)~\textbf{$\tau^2$-Bench}~\cite{tau2bench} is a multi-turn agent benchmark that simulates realistic customer service interactions across three domains: Retail, Airline, and Telecom. Each domain requires the agent to follow domain-specific policies and execute tool calls to resolve user requests. Since the training data only consists of the Retail domain, evaluation on Airline and Telecom can be regarded as a cross-domain scenario. (2)~\textbf{LiveCodeBench v6}~\cite{livecodebench} is a contamination-free code generation benchmark comprising problems of varying difficulty (Easy, Medium, Hard) sourced from recent programming contests. (3)~\textbf{AppWorld}~\cite{appworld} is an interactive agent benchmark that requires autonomous app usage through API calls across diverse everyday applications. We use AppWorld as an out-of-distribution evaluation to assess cross-task generalization, as it is entirely unseen during training.

\subsection{Main Results}
\begin{table*}[t]
\centering
\resizebox{\textwidth}{!}{%
\begin{tabular}{@{}llcccccccc@{}}
\toprule
\multicolumn{2}{c|}{\multirow{2}{*}{Method}} &
  \multicolumn{4}{c|}{Tau2Bench} &
  \multicolumn{4}{c}{LiveCodeBench v6} \\
\multicolumn{2}{c|}{} &
  Retail &
  Airline &
  Telecom &
  \multicolumn{1}{c|}{Overall} &
  Easy &
  Medium &
  Hard &
  Overall \\ \midrule

\multicolumn{10}{c}{\textit{\textbf{Skill Executor: Qwen3.5-4B}}} \\ \midrule
\multicolumn{2}{l|}{No Skill} &
  71.3 &
  68.8 &
  93.1 &
  \multicolumn{1}{c|}{79.5} &
  \textit{99.1} &
  67.3 &
  26.8 &
  56.6 \\
\multicolumn{2}{l|}{MemP} &
  72.5 &
  77.5 &
  93.8 &
  \multicolumn{1}{c|}{82.0} &
  \textbf{100.0} &
  42.3 &
  22.5 &
  47.4 \\ \midrule
\multirow{3}{*}{Qwen3.5-4B} &
  \multicolumn{1}{l|}{offline} &
  72.5 &
  76.3 &
  92.5 &
  \multicolumn{1}{c|}{81.3} &
  97.7 &
  61.5 &
  25.0 &
  53.7 \\
 &
  \multicolumn{1}{l|}{online} &
  70.6 &
  \textit{80.0} &
  \textit{96.3} &
  \multicolumn{1}{c|}{82.8} &
  98.1 &
  68.9 &
  26.5 &
  56.7 \\
 &
  \multicolumn{1}{l|}{hybrid} &
  73.8 &
  77.5 &
  93.8 &
  \multicolumn{1}{c|}{82.5} &
  95.4 &
  64.1 &
  29.6 &
  56.0 \\ \midrule
\multirow{3}{*}{SkillComposer-4B} &
  \multicolumn{1}{l|}{offline} &
  72.5 &
  75.0 &
  92.5 &
  \multicolumn{1}{c|}{81.0} &
  95.4 &
  65.4 &
  25.8 &
  54.7 \\
 &
  \multicolumn{1}{l|}{online} &
  \textit{74.4} &
  78.8 &
  \textbf{98.1} &
  \multicolumn{1}{c|}{84.8} &
  98.1 &
  \textbf{72.7} &
  \textit{29.3} &
  \textbf{59.1} \\
 &
  \multicolumn{1}{l|}{hybrid} &
  \textbf{77.5} &
  \textbf{82.5} &
  95.6 &
  \multicolumn{1}{c|}{\textbf{85.7}} &
  98.5 &
  \textit{69.2} &
  \textbf{31.3} &
  \textit{59.0} \\ \midrule

\multicolumn{10}{c}{\textit{\textbf{Skill Executor: Qwen3.5-27B}}} \\ \midrule
\multicolumn{2}{l|}{No Skill} &
  75.6 &
  77.5 &
  94.4 &
  \multicolumn{1}{c|}{83.5} &
  99.5 &
  90.4 &
  62.0 &
  79.7 \\
\multicolumn{2}{l|}{MemP} &
  71.3 &
  78.8 &
  97.5 &
  \multicolumn{1}{c|}{83.3} &
  \textbf{100.0} &
  67.3 &
  47.5 &
  66.3 \\ \midrule
\multirow{3}{*}{Qwen3.5-4B} &
  \multicolumn{1}{l|}{offline} &
  72.5 &
  73.8 &
  96.3 &
  \multicolumn{1}{c|}{82.3} &
  \textbf{100.0} &
  84.6 &
  55.0 &
  74.9 \\
 &
  \multicolumn{1}{l|}{online} &
  77.5 &
  78.8 &
  \textbf{98.8} &
  \multicolumn{1}{c|}{86.3} &
  \textbf{100.0} &
  91.9 &
  64.5 &
  81.4 \\
 &
  \multicolumn{1}{l|}{hybrid} &
  77.5 &
  80.0 &
  97.5 &
  \multicolumn{1}{c|}{86.0} &
  99.2 &
  90.4 &
  64.2 &
  80.6 \\ \midrule
\multirow{3}{*}{SkillComposer-4B} &
  \multicolumn{1}{l|}{offline} &
  73.8 &
  73.8 &
  97.5 &
  \multicolumn{1}{c|}{83.3} &
  98.3 &
  89.4 &
  59.7 &
  78.0 \\
 &
  \multicolumn{1}{l|}{online} &
  \textbf{79.4} &
  \textbf{83.5} &
  \textbf{98.8} &
  \multicolumn{1}{c|}{\textbf{88.0}} &
  \textbf{100.0} &
  \textbf{94.2} &
  \textit{66.8} &
  \textbf{83.1} \\
 &
  \multicolumn{1}{l|}{hybrid} &
  \textit{78.1} &
  \textit{81.3} &
  \textbf{98.8} &
  \multicolumn{1}{c|}{\textit{87.0}} &
  99.4 &
  \textit{93.3} &
  \textbf{67.2} &
  \textit{82.9} \\ \bottomrule
\end{tabular}%
}
\caption{Main results on $\tau^2$-Bench (agent) and LiveCodeBench v6 (code). Rejection sampling and offline library construction use Qwen3.5-4B as the skill executor, with the $\tau^2$-Bench training set and randomly sampled 500 OpenCoderReasoning samples as data sources. \textbf{Bold} denotes the best and \textit{italic} the second best within each executor.}
\label{tab:main-results}
\end{table*}
Table~\ref{tab:main-results} presents results on $\tau^2$-Bench and LiveCodeBench v6. We compare against \textit{No Skill} (vanilla LLM) and \textit{MemP}~\cite{memp} (procedural memory extracted from valid trajectories), and additionally report results for the untrained Qwen3.5-4B to isolate the contribution of rejection sampling SFT. All training data is collected using the 4B model. For offline skill library construction, we use the training set of $\tau^2$-Bench for agent tasks and 500 randomly sampled instances from OpenCoderReasoning for code tasks.

\paragraph{Effectiveness of \method{}.}
On the 4B executor, \method{} consistently outperforms both baselines and its untrained counterparts across all modes. The hybrid mode achieves the best $\tau^2$-Bench score (85.7, $+$6.2 over No Skill), while the online mode leads on LiveCodeBench (59.1, $+$2.5). MemP degrades LiveCodeBench performance to 47.4, well below the No Skill baseline of 56.6, underscoring the risk of uncontrolled skill extraction. In contrast, \method{} online improves over its untrained counterpart by 2.0 points on $\tau^2$-Bench and 2.4 points on LiveCodeBench, confirming that rejection-sampling-based training yields meaningful quality gains. We provide detailed performance curves across iterations in Appendix~\ref{sec:iteration-curves} (Figure~\ref{fig:iteration-curves}).

Moreover, we further provide inference efficiency comparison across the base model and our trained SkillComposer-4B. As shown in table~\ref{fig:acc_tokens_trend}, as skills evolve, the average token consumption of both methods decreases while the average performance keep increasing, which indicates the steady performance and efficiency of our methods. Notably, except for the performance superiority, our trained SkillComposer-4B also consistently show better efficiency.

\paragraph{Comparison across modes.}
Online and hybrid consistently outperform offline, indicating that iterative skill refinement at inference time provides substantial gains beyond static retrieval. Hybrid achieves the strongest $\tau^2$-Bench results by combining a generalized library with online specialization. Notably, the performance gains of offline and hybrid modes are more pronounced on $\tau^2$-Bench than on LiveCodeBench. We attribute this to the offline library construction: the $\tau^2$-Bench library is built from its in-distribution training set, while the code library draws from only 500 randomly sampled OpenCoderReasoning instances, which offers limited coverage of LiveCodeBench's task distribution. 

In contrast, the online mode, which requires no pre-built library and instead refines skills per task, excels on LiveCodeBench, where task-specific specification is most beneficial. Furthermore, on Hard-level code problems, hybrid consistently outperforms online (31.3 vs. 29.3 on 4B; 67.2 vs. 66.8 on 27B), suggesting that an initial skill library serves as valuable prior experience that particularly benefits the executor on more challenging tasks where structured guidance is most needed. However, on the Simple and Medium-Level problems, hybrid lags behind the online setting, which indicates that unnecessary skills might mislead the language models when facing simpler tasks.

\subsection{Cross-Model Generalization}

Since all training data is collected with Qwen3.5-4B as the skill executor, the 27B results directly evaluate cross-model transfer. As shown in Table~\ref{tab:main-results}, \method{} generalizes effectively across model scales: SkillComposer-4B-online achieves 88.0 on $\tau^2$-Bench ($+$4.5 over No Skill) and 83.1 on LiveCodeBench ($+$3.4), while the hybrid mode reaches 67.2 on Hard-level code problems. These results confirm that skills composed by a smaller model remain effective when consumed by a substantially larger executor, suggesting that skill quality is largely decoupled from the composer's capacity.

\subsection{Cross-Domain Generalization}

Our $\tau^2$-Bench training data is sourced exclusively from the Retail domain, making Airline and Telecom results a direct test of cross-domain generalization. On the 4B executor, SkillComposer-4B-hybrid achieves 82.5 on Airline ($+$13.7 over No Skill) and 95.6 on Telecom ($+$2.5). On the 27B executor, SkillComposer-4B-online reaches 83.5 on Airline ($+$6.0) and 98.8 on Telecom ($+$4.4). The Airline gains suggest that skills learned from Retail, which shares structural similarities as a customer-service domain, transfer effectively to related but unseen domains. This indicates that \method{} acquires generalizable composition abilities rather than domain-specific patterns.

\section{Ablation and Analysis}

We now investigate the mechanisms behind \method{}'s gains. We first ablate the individual skill operations to verify that merge and improve serve complementary roles (\S\ref{subsec:ablation-ops}). We then ask whether the learned composition ability is tied to the training domain or transfers across task types (\S\ref{subsec:transfer}), whether it generalizes to entirely unseen benchmarks (\S\ref{subsec:cross-task}), and whether iterative skill evolution offers genuine advantages over brute-force repeated sampling (\S\ref{subsec:evolution-vs-sampling}).

\subsection{Ablation on Skill Operations}
\label{subsec:ablation-ops}
We ablate the skill operations used during offline library construction to isolate each operation's contribution. Table~\ref{tab:skill-ops} reports offline-mode results with Qwen3.5-4B as the skill executor.

\begin{table}[htbp]
\centering
\resizebox{\columnwidth}{!}{%
\begin{tabular}{@{}l|cc@{}}
\toprule
\multicolumn{1}{c|}{\textbf{Skill   Operations}} & \textbf{$\tau^2$-Bench} & \textbf{LiveCodeBench v6} \\ \midrule
Create               & 73.8          & \textit{58.5} \\
Create/Improve       & 75.6          & 57.3          \\
Create/Merge         & \textbf{77.5} & \textbf{59.0} \\
Create/Merge/Improve & \textit{76.9} & 57.9          \\ \bottomrule
\end{tabular}%
}
\caption{Ablation on skill operations used during offline library construction (Qwen3.5-4B executor). Skill Merge provides the largest gain by promoting generalization, while Skill Improve is less beneficial in the offline setting, where transferability is prioritized.}
\label{tab:skill-ops}
\end{table}

\begin{figure}[t]
    \centering
    \includegraphics[width=\columnwidth]{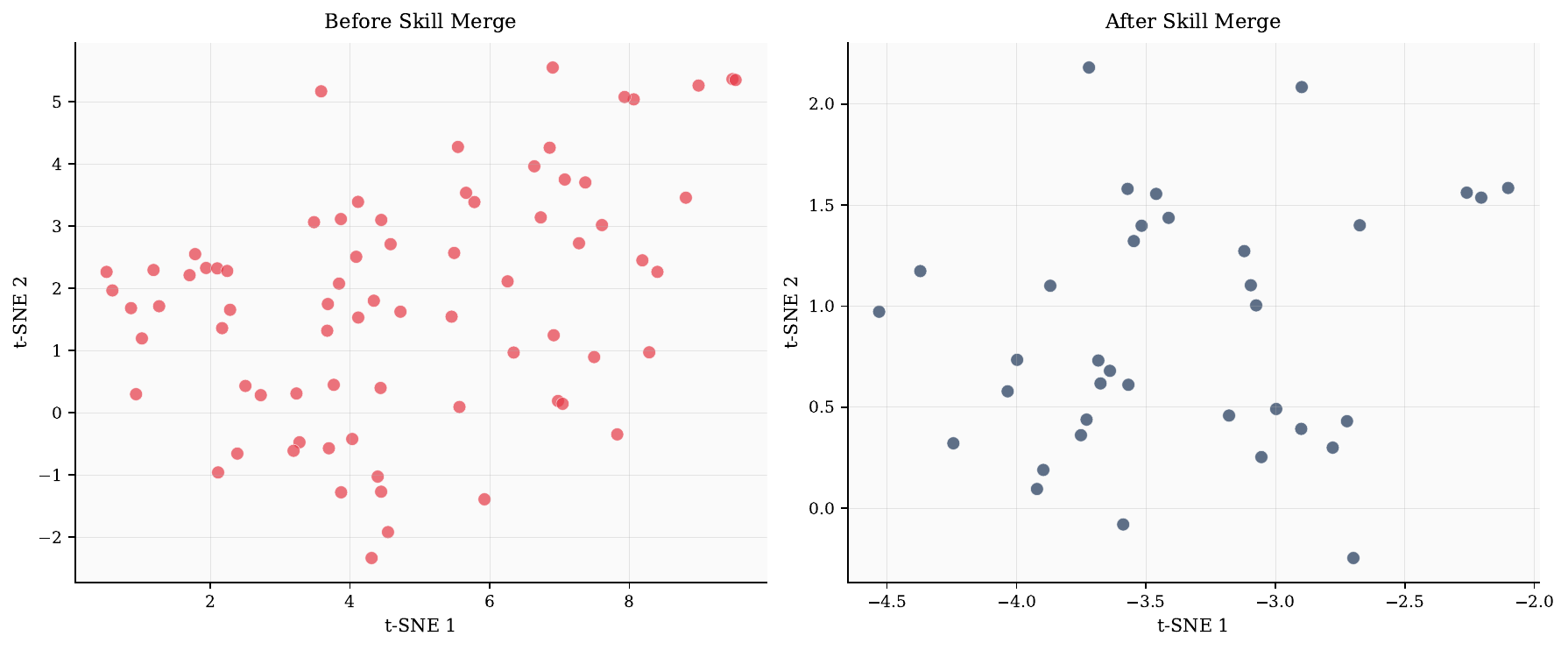}
    \caption{t-SNE visualization of the Retail skill library before (left) and after (right) Skill Merge. Merging consolidates semantically similar skills, producing a \textbf{sparser} and \textbf{more general} library with reduced redundancy.}
    \label{fig:tsne}
\end{figure}

Skill Merge is the most impactful operation: adding it to Skill Create yields $+$3.7 on $\tau^2$-Bench and $+$0.5 on LiveCodeBench. This aligns with our framework's design, as offline libraries prioritize generalized skills, and Skill Merge is precisely the operation that drives generalization. To visualize this effect, Figure~\ref{fig:tsne} shows t-SNE embeddings of the Retail skill library before and after merging. After Skill Merge, the skill distribution becomes notably sparser, confirming that merge effectively consolidates redundant skills into fewer, more general ones. Adding Skill Improve alone to Create provides moderate gains on $\tau^2$-Bench ($+$1.8) but slightly hurts LiveCodeBench ($-$1.2), suggesting that specification can reduce transferability for unseen tasks. Combining all three operations performs slightly below Skill Create/Merge, indicating that indiscriminate specification during offline construction partially counteracts the generalization benefit of Skill Merge. These results reinforce that Skill Improve is better suited for online and hybrid settings, where skills are refined toward specific target tasks.

\subsection{Transferability of Skill Composition}
\label{subsec:transfer}
We investigate whether the skill composition ability generalizes across task types or remains specific to the training scenario. We train \method{} with agent-only, code-only, and mixed (agent \& code) rejection sampling data, evaluating all variants on LiveCodeBench v6 in online mode with Qwen3.5-27B as the skill executor (Table~\ref{tab:cross-scenario}).

\begin{table}[htbp]
\centering
\resizebox{\columnwidth}{!}{%
\begin{tabular}{@{}l|ccc|c@{}}
\toprule
\multicolumn{1}{c|}{SkillComposer} & \multicolumn{1}{c}{Easy} & \multicolumn{1}{c}{Medium} & \multicolumn{1}{c|}{Hard} & \multicolumn{1}{c}{Overall} \\ \midrule
\multicolumn{1}{l|}{Qwen3.5-4B} & \textbf{100.0} & 91.9          & 64.5          & 81.4          \\
~~~~~~~~~~+agent                          & \textbf{100.0} & 93.5          & \textit{66.8} & 82.9          \\
~~~~~~~~~~+code                           & 99.5           & \textbf{95.8} & \textbf{68.0} & \textbf{84.0} \\
~~~~~~~~~~+agent \& code                  & \textbf{100.0} & \textit{94.2} & \textit{66.8} & \textit{83.1} \\ \bottomrule
\end{tabular}%
}
\caption{Transferability of skill composition across task types, evaluated on LiveCodeBench v6 (online mode, Qwen3.5-27B executor). Agent-only training still improves code performance, confirming that skill composition is a general-purpose ability.}
\label{tab:cross-scenario}
\end{table}

Training with agent-only data improves overall performance from 81.4 to 82.9 ($+$1.5), despite never encountering code tasks during training. This confirms that skill composition is a general-purpose ability: structuring procedural knowledge and iterative refinement transfer across task types. Code-only training yields the strongest results (84.0, $+$2.6) due to direct domain alignment, while the mixed setting used in practice achieves 83.1, offering a balanced trade-off.

\subsection{Cross-Task Generalization}
\label{subsec:cross-task}
To further evaluate whether \method{} generalizes to entirely unseen agent tasks, we test the online mode on AppWorld~\cite{appworld}, a benchmark not used during training or rejection sampling. We compare the vanilla model (No Skill), the untrained Qwen3.5-4B composer, and SkillComposer-4B, all using Qwen3.5-4B as the skill executor. Results are shown in Table~\ref{tab:appworld}.

SkillComposer-4B achieves the best performance on both metrics, improving accuracy to 57.7 ($+$2.5 over Vanilla) while maintaining a high pass rate of 88.3. Notably, the untrained composer improves accuracy (57.1) but degrades pass rate from 88.1 to 86.0, indicating that naively generated skills can introduce errors that reduce task completion reliability. In contrast, \method{} improves both metrics simultaneously, confirming that rejection-sampling-based training produces skills that enhance correctness without sacrificing robustness. This result demonstrates that the skill composition ability learned from $\tau^2$-Bench transfers effectively to a structurally different agent benchmark.

\begin{table}[htbp]
\centering
\resizebox{0.9\columnwidth}{!}{%
\begin{tabular}{@{}c|ccc@{}}
\toprule
Method    & Vanilla & Qwen3.5-4B & SkillComposer-4B \\ \midrule
Accuracy  & 55.2    & 57.1       & \textbf{57.7}    \\
Pass Rate & 88.1    & 86.0       & \textbf{88.3}    \\ \bottomrule
\end{tabular}%
}
\caption{Cross-task generalization on AppWorld (online mode, Qwen3.5-4B executor). \method{} improves both accuracy and pass rate, while the untrained composer degrades pass rate.}
\label{tab:appworld}
\end{table}

\subsection{Iterative Evolution vs. Repeated Sampling}
\label{subsec:evolution-vs-sampling}
A natural question is whether iterative skill evolution offers advantages beyond simply sampling more attempts. To answer this, we compare two strategies under the same inference budget of $k$ rounds: \textit{i}) vanilla pass@$k$, which independently samples $k$ solutions without any skill and selects the best one, and \textit{ii}) \method{} online evolution, which creates a skill from the first trajectory and iteratively improves it across $k$ rounds.

\begin{figure}[t]
    \centering
    \includegraphics[width=\columnwidth]{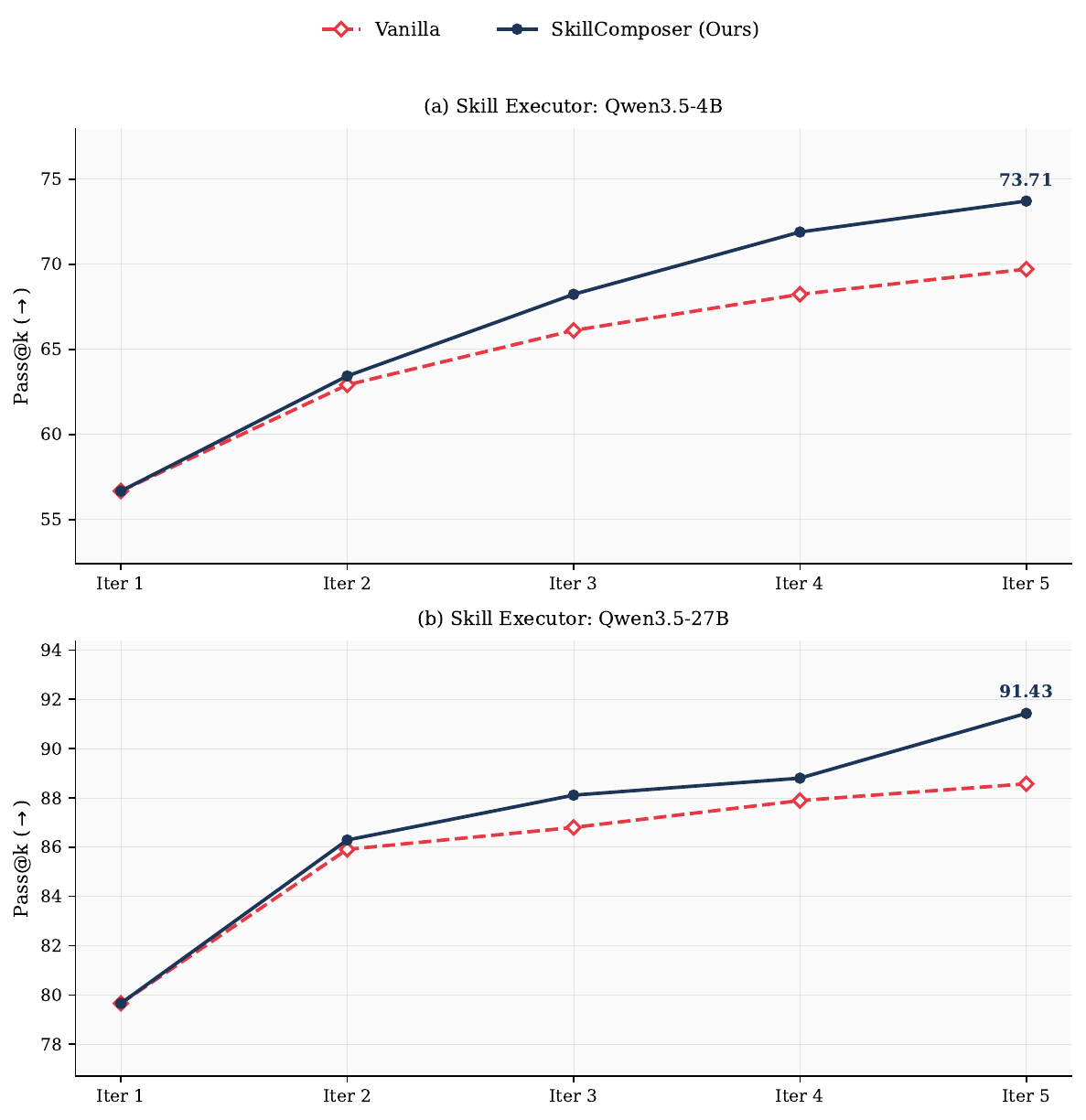}
    \caption{Pass@$k$ vs. \method{} online evolution over $k$ rounds on LiveCodeBench v6. Both strategies share the same starting point at $k{=}1$. Iterative skill refinement consistently outperforms independent repeated sampling, with the gap widening as $k$ increases.}
    \label{fig:pass-at-k-by-round}
\end{figure}

As shown in Figure~\ref{fig:pass-at-k-by-round}, both strategies start from the same baseline at $k{=}1$, but \method{} consistently outperforms vanilla pass@$k$ as the number of rounds increases. On the 4B executor, \method{} reaches 73.71 at $k{=}5$ compared to 69.71 for vanilla pass@$k$, a gain of 4.0 points. On the 27B executor, the gap persists at 91.43 vs. 88.57 ($+$2.86). Notably, the performance gap widens with each additional round, indicating that iterative skill refinement compounds its advantage over independent sampling. This demonstrates that structured skill evolution is a more effective use of additional inference budget than brute-force repeated sampling.

\section{Conclusion}

We presented \method{}, a framework that decomposes skill construction into three learnable operations: create, improve, and merge. By training these operations via rejection sampling filtered by delta pass rate, \method{} enables language models to self-evolve skill libraries at inference time without ground-truth supervision. The framework supports three complementary modes: offline for building generalized libraries, online for task-specific refinement, and hybrid for combining both. Experiments on $\tau^2$-Bench, LiveCodeBench v6, and AppWorld demonstrate that \method{} consistently outperforms baselines and generalizes across model scales, domains, and task types. Our analysis further reveals that merge drives generalization while improve drives specification, and that skill composition is a transferable ability that benefits from but does not require in-domain training data.

\section*{Limitations}

The rejection sampling procedure requires multiple inference passes per task to compute delta pass@$k$, making training data collection computationally expensive. Additionally, our experiments are conducted with a limited number of model families (Qwen3.5-4B and Qwen3.5-27B); evaluating \method{} across a broader range of architectures and scales would strengthen the generality of our findings.



\bibliography{main}


\newpage
\appendix
\section{Implementation Details}
\label{app:implementation}
In this paper, we use Qwen3-Embedding-8B~\cite{qwen3embedding} as the embedding model to calculate the similarity. We list the global hyperparameter settings across all tasks below:
\begin{table}[htbp]
    \begin{center}
        \label{tab:eval-details}
        \resizebox{0.8\columnwidth}{!}{%
            \begin{tabular}{@{}l|l@{}}
                \toprule
                \textbf{Parameter}          & \textbf{Value} \\ \midrule
                similarity threshold $\delta$ & 0.8           \\
                delta pass rate for Skill Create $\epsilon_{\rm c}$ & 0.4\\
                delta pass rate for Skill Merge $\epsilon_{\rm m}$ & 0.4\\
                delta pass rate for Skill Improve $\epsilon_{\rm i}$ & 0.4\\\bottomrule
            \end{tabular}
        }
        \caption{Hyperparameters details for rejection sampling.}
    \end{center}
\end{table}
\subsection{Code}
\paragraph{Rejection Sampling} For rejection sampling, we use 8,000 data points sampled from OpenCodeReasoning, and run 5 separate runs for each data point.

\paragraph{Skill Library Building} For the initial skill library building of offline and hybrid modes, we incorporate 500 randomly sampled data points from the OpenCodeReasoning dataset.

\subsection{Agent}

\paragraph{Skill Library Building} For the initial skill library building of offline and hybrid modes, we use the training split of $\tau^2$-Bench.

\subsection{Evaluation Details}
For agent tasks, we conduct 4 separate runs across all methods and settings. For code tasks, we conduct 5 separate runs across all methods and settings. Finally, we report the average score of the best iteration across all runs for each setting. Moreover, during the skill library building of the evaluation process, we use the same similarity threshold $\delta$ as rejection sampling.

\subsection{Training Details}
For supervised fine-tuning, we trained Qwen3.5-4B with LlamaFactory. We list the detailed hyperparameters below:
\begin{table}[htbp]
    \begin{center}
        \label{tab:sft-details}
        \resizebox{0.7\columnwidth}{!}{%
            \begin{tabular}{@{}l|l@{}}
                \toprule
                \textbf{Parameter}          & \textbf{Value} \\ \midrule
                Learning Rate               & 1e-6           \\
                Train Batch Size            & 64             \\
                Number of Training Epochs   & 3              \\
                Max Length                  & 65536              \\
                LR Scheduler                & cosine         \\
                Warmup Ratio                & 0.1            \\\bottomrule
            \end{tabular}
        }
        \caption{Supervised fine-tuning details for \method{}.}
    \end{center}
\end{table}

\section{Rejection Sampling Procedures}
\label{sec:rejection-sampling-appendix}

We provide detailed pseudo-code for the three rejection sampling procedures introduced in \S\ref{subsec:rejection-sampling}.

\begin{algorithm}[H]
\caption{Rejection Sampling for Skill Create}
\label{alg:rs-create}
\begin{algorithmic}[1]
\REQUIRE Task set $\mathcal{X}$, skill executor LLM, base composer $\mathcal{C}$, samples $n$, threshold $\epsilon_{\rm c}$
\ENSURE Training set $\mathcal{D}_{\text{create}}$
\STATE $\mathcal{D}_{\text{create}} \leftarrow \emptyset$
\FOR{each task $x \in \mathcal{X}$}
    \STATE $p_{\text{base}} \leftarrow \text{pass@}1(\text{LLM}(x), n)$
    \STATE $\tau \leftarrow$ sample one trajectory from LLM$(x)$
    \STATE $s \leftarrow \mathcal{C}_{\text{create}}(x, \tau)$
    \STATE $p_{\text{skill}} \leftarrow \text{pass@}1(\text{LLM}(x, s), n)$
    \IF{$p_{\text{skill}} - p_{\text{base}} \geq \epsilon_{\rm c}$}
        \STATE $\mathcal{D}_{\text{create}} \leftarrow \mathcal{D}_{\text{create}} \cup \{(x, \tau) \rightarrow s\}$
    \ENDIF
\ENDFOR
\end{algorithmic}
\end{algorithm}

\begin{algorithm}[H]
\caption{Rejection Sampling for Skill Merge}
\label{alg:rs-merge}
\begin{algorithmic}[1]
\REQUIRE Skill-task pairs $\{(s_i, x_i)\}$, skill executor LLM, base composer $\mathcal{C}$, samples $n$, threshold $\epsilon_{\rm m}$
\ENSURE Training set $\mathcal{D}_{\text{merge}}$
\STATE $\mathcal{D}_{\text{merge}} \leftarrow \emptyset$
\STATE Compute pairwise similarity among all skills
\FOR{each similar pair $(s_1, x_1)$, $(s_2, x_2)$}
    \STATE $s \leftarrow \mathcal{C}_{\text{merge}}(s_1, s_2)$
    \STATE $\Delta \leftarrow \frac{1}{2}\sum_{i=1}^{2}[\text{pass@}1(\text{LLM}(x_i, s), n) - \text{pass@}1(\text{LLM}(x_i, s_i), n)]$
    \IF{$\Delta \geq \epsilon_{\rm m}$}
        \STATE $\mathcal{D}_{\text{merge}} \leftarrow \mathcal{D}_{\text{merge}} \cup \{(s_1, s_2) \rightarrow s\}$
    \ENDIF
\ENDFOR
\end{algorithmic}
\end{algorithm}

\begin{algorithm}[H]
\caption{Rejection Sampling for Skill Improve}
\label{alg:rs-improve}
\begin{algorithmic}[1]
\REQUIRE Skill-task pairs $\{(s_i, x_i)\}$, skill executor LLM, base composer $\mathcal{C}$, samples $n$, threshold $\epsilon_{\rm i}$
\ENSURE Training set $\mathcal{D}_{\text{improve}}$
\STATE $\mathcal{D}_{\text{improve}} \leftarrow \emptyset$
\STATE Compute skill-task similarity matrix
\FOR{each selected pair $(s_{\rm o}, x')$ with original task $x$}
    \STATE $\tau' \leftarrow$ sample one trajectory from LLM$(x', s_{\rm o})$
    \STATE $s \leftarrow \mathcal{C}_{\text{improve}}(x', s_{\rm o}, \tau')$
    \STATE $\Delta \leftarrow \frac{1}{2}\sum_{x_i \in \{x, x'\}}[\text{pass@}1(\text{LLM}(x_i, s), n) - \text{pass@}1(\text{LLM}(x_i, s_{\rm o}), n)]$
    \IF{$\Delta \geq \epsilon_{\rm i}$}
        \STATE $\mathcal{D}_{\text{improve}} \leftarrow \mathcal{D}_{\text{improve}} \cup \{(x', s_{\rm o}, \tau') \rightarrow s\}$
    \ENDIF
\ENDFOR
\end{algorithmic}
\end{algorithm}

\section{Performance by Iteration}
\label{sec:iteration-curves}

\begin{figure*}[htbp]
    \centering
    \includegraphics[width=\textwidth]{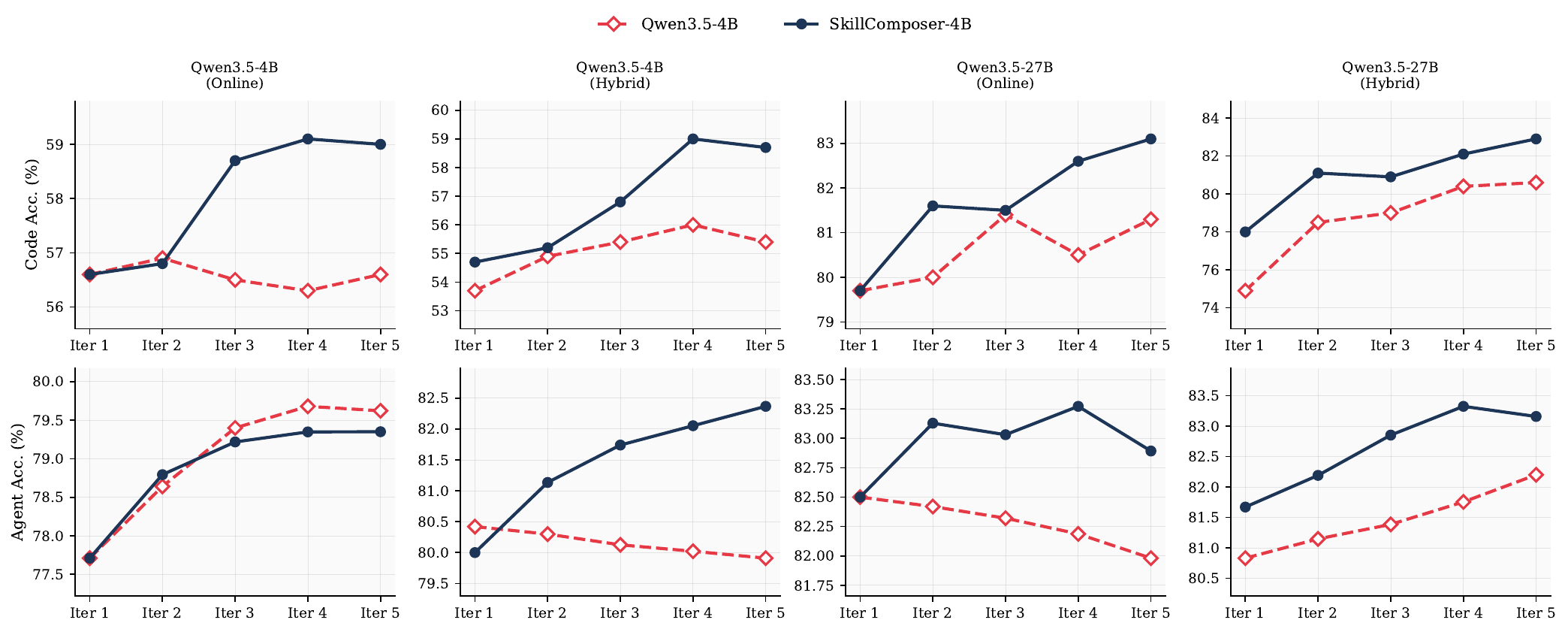}
    \caption{Average performance by iteration for online and hybrid modes on $\tau^2$-Bench (average scores across Retail, Airline, Telecom) and LiveCodeBench v6. \method{} exhibits steeper and more sustained improvement across iterations compared to the untrained baselines in both settings.}
    \label{fig:iteration-curves}
\end{figure*}

Figure~\ref{fig:iteration-curves} presents the performance trajectory across iterations for both online and hybrid modes. For $\tau^2$-Bench, we report the overall score across all three domains (Retail, Airline, and Telecom). The curves illustrate how skill quality evolves over successive iterations, with \method{} consistently achieving steeper improvement curves compared to the untrained model in both settings.

\section{Inference Efficiency across Iterations}
\begin{figure*}[t]
    \centering
    \includegraphics[width=\textwidth]{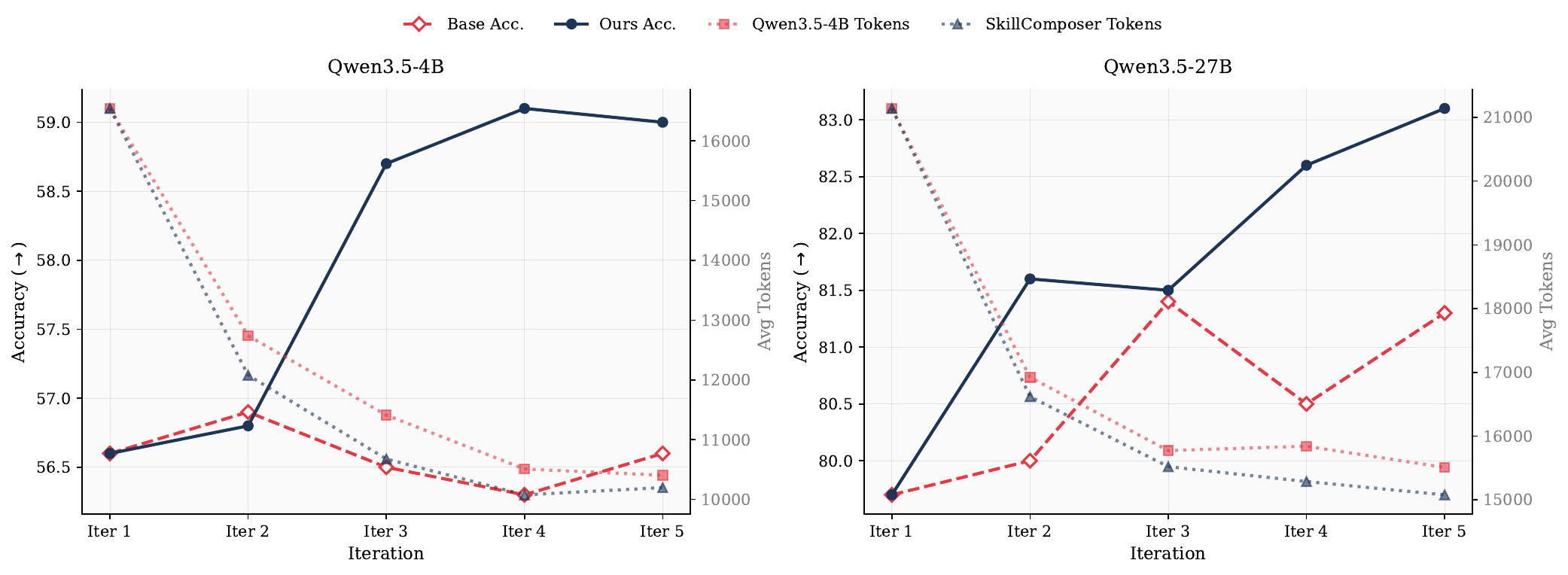}
    \caption{The token consumption across iterations of Qwen3.5-4B and SkillComposer-4B.}
    \label{fig:acc_tokens_trend}
\end{figure*}

\section{Prompt Template}
\label{sec:prompt}

We present the prompt templates used for the four core operations in \method{}. Each prompt is instantiated with task-specific variables (denoted by \texttt{\{\{variable\}\}}) at inference time.

\subsection{Skill Create Prompt}

The following prompt instructs the skill composer to extract a reusable skill from a completed task trajectory.

\begin{tcolorbox}[colback=gray!5, colframe=gray!50, breakable, title=Skill Create Prompt]
\small
\texttt{You are a professional skill creator. A skill is an executable procedure that provides specialized instructions for specific tasks.}

\vspace{0.5em}
\texttt{[Task Description]}\\
\texttt{1. Given a task and a corresponding trajectory generated by a language model, your task is to generate a reusable/generalizable skill based on the given contents.\\
2. The reusability/generalizability should be both model-level and task-level:\\
  - The skill should help current language model not only solve current task but this type of tasks\\
  - The skill should help other language models not only solve current task but this type of tasks}

\vspace{0.5em}
\texttt{[Skill Formulation]}\\
\texttt{The generated skill should consist of three components including name, description and body:\\
1. **name**: The identifier of the skill. Once the name is generated, it will never be further modified. Therefore, make its name as general and abstract as possible, rather than limiting it to the current input problem and trajectory.\\
2. **description**: When to trigger, what it does. This is the primary triggering mechanism - include both what the skill does AND specific contexts for when to use it. All "when to use" info goes here, not in the body.\\
3. **body**: A document in markdown-style. It should contain the name of the skill and several sections with detailed instructions (i.e., what you want the skill to do and roughly how it should do it) to solve this type of tasks.}

\vspace{0.5em}
\texttt{[Skill Create Instructions]}\\
\texttt{1. When writing a skill, try to explain to the model why things are important in lieu of heavy-handed musty MUSTs.\\
2. Use theory of mind and try to make the skill general and not super-narrow to specific examples.\\
3. Start by writing a draft and then look at it with fresh eyes and improve it.\\
4. Make sure not to introduce any specific concepts, examples or details from the input.}

\vspace{0.5em}
\texttt{[Input Details]}\\
\texttt{Here are the task and the trajectory:\\\#\#\# Task\\\{\{task\}\}\\\#\#\# Trajectory\\\{\{trajectory\}\}}

\vspace{0.5em}
\texttt{[Output Format]}\\
\texttt{Put your final output in a JSON object enclosed by a json code block like:\\```json\\\{\{\\"name": <skill-name>,\\"description": <skill-description>,\\"body": <skill-body>\\\}\}\\```}
\end{tcolorbox}

\subsection{Skill Merge Prompt}

The following prompt instructs the skill composer to merge two related skills into a more general one.

\begin{tcolorbox}[colback=gray!5, colframe=gray!50, breakable, title=Skill Merge Prompt]
\small
\texttt{You are a professional Skill Merger. A skill is an executable procedure that provides specialized instructions for specific tasks.}

\vspace{0.5em}
\texttt{[Task Description]}\\
\texttt{1. Given two seperate skills, your task is to merge them into a more generalizable and reusable skill.\\
2. The merged skill should cover both the common and diverse instructions within the given two skills. The reusability/generalizability should be both model-level and task-level:\\
  - The skill should help current language model not only solve current question but this type of questions\\
  - The skill should help other language models not only solve current question but this type of questions}

\vspace{0.5em}
\texttt{[Skill Formulation]}\\
\texttt{The generated skill should consist of three components including name, description and body:\\
1. **name**: The identifier of the skill. Once the name is generated, it will never be further modified. Therefore, make its name as general and abstract as possible, rather than limiting it to the current input problem and trajectory.\\
2. **description**: When to trigger, what it does. This is the primary triggering mechanism - include both what the skill does AND specific contexts for when to use it. All "when to use" info goes here, not in the body.\\
3. **body**: A document in markdown-style. It should contain the name of the skill and several sections with detailed instructions (i.e., what you want the skill to do and roughly how it should do it) to solve this type of tasks.}

\vspace{0.5em}
\texttt{[Skill Merge Instructions]}\\
\texttt{1. When merging two skills, try to explain to the model why things are important in lieu of heavy-handed musty MUSTs.\\
2. Use theory of mind and try to make the skill general and not super-narrow to specific examples.\\
3. The merged skill should cover both the common and diverse instructions within the given two skills.\\
4. Merge the two skills into one standalone, executable skill, not a simple combination.\\
5. Remove overlap and unify similar instructions into the clearest version.\\
6. Make the final result clearer, cleaner, more reusable and more generalizable than either original skill alone.}

\vspace{0.5em}
\texttt{[Input Details]}\\
\texttt{\#\#\# Skill A\\---\\name: \{name\_a\}\\description: \{description\_a\}\\---\\\{body\_a\}\\\\\#\#\# Skill B\\---\\name: \{name\_b\}\\description: \{description\_b\}\\---\\\{body\_b\}}

\vspace{0.5em}
\texttt{[Output Format]}\\
\texttt{Put your final output in a JSON object enclosed by a json code block like:\\```json\\\{\{\\"name": <skill-name>,\\"description": <skill-description>,\\"body": <skill-body>\\\}\}\\```}
\end{tcolorbox}

\subsection{Skill Improve Prompt}

The following prompt instructs the skill composer to refine an existing skill based on a new task trajectory.

\begin{tcolorbox}[colback=gray!5, colframe=gray!50, breakable, title=Skill Improve Prompt]
\small
\texttt{You are a professional Skill Improver. A skill is an executable procedure that provides specialized instructions for specific tasks.}

\vspace{0.5em}
\texttt{[Task Description]}\\
\texttt{1. Given a question, an original skill, and a corresponding trajectory generated by a language model when injecting this original skill into its context, your task is to generate an improved version of the skill.\\
2. The improved skill should maintain its reusability/generalizability at both model-level and task-level:\\
  - The skill should help current language model not only solve current question but this type of questions\\
  - The skill should help other language models not only solve current question but this type of questions}

\texttt{[Expert Analysis Before Improvement]}\\
\texttt{Before designing the improved skill, you must act as an expert evaluator:\\
- Carefully examine the trajectory and judge whether it fully completed the user's request.\\
- Decompose the trajectory's completion degree: identify specific parts that were executed well and parts that were insufficient or incorrect.\\
- If the request was completed successfully, extract reliable, generalizable experiences (best practices) that likely contributed to the success, and check whether the current skill already captures them.\\
- If certain aspects were not handled well, summarize the lessons learned and points of attention that the improved skill should explicitly warn about or guide the model to avoid.\\
- Use this analysis to inform the improved skill, ensuring that the final skill encodes both proven strategies and preventive guidance.}

\vspace{0.5em}
\texttt{[Skill Formulation]}\\
\texttt{The generated skill should consist of three components including name, description and body:\\
1. **name**: The identifier of the skill. Once the name is generated, it will never be further modified. Therefore, make its name as general and abstract as possible, rather than limiting it to the current input problem and trajectory.\\
2. **description**: When to trigger, what it does. This is the primary triggering mechanism - include both what the skill does AND specific contexts for when to use it. All "when to use" info goes here, not in the body.\\
3. **body**: A document in markdown-style. It should contain the name of the skill and several sections with detailed instructions (i.e., what you want the skill to do and roughly how it should do it) to solve this type of tasks.}

\vspace{0.5em}
\texttt{[Skill Improve Instructions]}\\
\texttt{1. Use the original skill as a foundation and produce an improved version that is clearer, more complete, and more reusable.\\
2. Preserve what already works, strengthen weak or missing guidance, and reflect the lessons from the trajectory.\\
3. Keep the skill generalizable across similar tasks rather than overfitting to the current example.\\
4. If the trajectory does not provide useful improvement signal, return the original skill unchanged.}

\vspace{0.5em}
\texttt{[Input Details]}\\
\texttt{Here are the question, the trajectory, and the original skill:\\\#\#\# Question\\\{\{question\}\}\\\#\#\# Trajectory\\\{\{trajectory\}\}\\\#\#\# Skill\\---\\name: \{name\}\\description: \{description\}\\---\\\{body\}}

\vspace{0.5em}
\texttt{[Output Format]}\\
\texttt{Put your final output in a JSON object enclosed by a json code block like:\\```json\\\{\{\\"name": <skill-name>,\\"description": <skill-description>,\\"body": <skill-body>\\\}\}\\```}
\end{tcolorbox}

\subsection{Skill Use System Prompt}

The following system prompt equips the skill executor with a skill to guide task solving.

\begin{tcolorbox}[colback=gray!5, colframe=gray!50, breakable, title=Skill Use Prompt]
\small
\texttt{You are a professional skill user. The following is a skill that provides specialized instructions for specific tasks.}

\texttt{name: \{name\}\\---\\\{body\}}

\texttt{Read the skill above carefully, and use its instructions to solve problems. If you think the skill is unuseful or even wrong, ignore it and use your own knowledge instead.}
\end{tcolorbox}

\begin{figure*}[htbp]
    \centering
    \includegraphics[width=\textwidth]{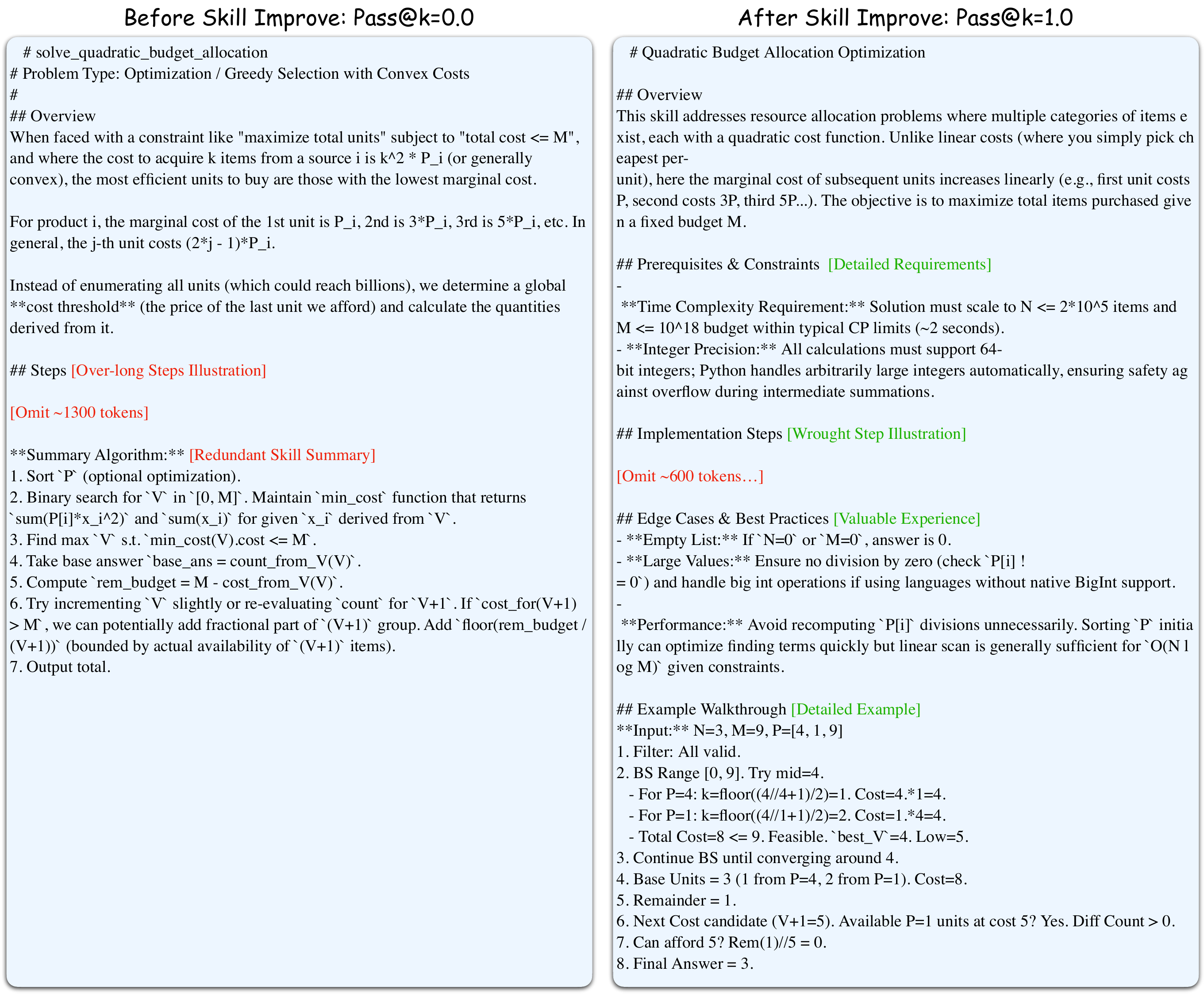}
    \caption{A case study on under-improved skill (left) and improved skill (right).}
    \label{fig:case}
\end{figure*}

\section{Case Studies}
Figure~\ref{fig:case} gives a case study on how skill evolves before and after Skill Improve. With the improved skill, the skill executor get higher pass@$k$ on the same task.

\end{document}